\documentclass{article}

\usepackage{arxiv}
\usepackage{hyperref}       
\usepackage{booktabs}       
\usepackage{amsfonts}       
\usepackage{graphicx}
\usepackage[numbers]{natbib}
\usepackage{subcaption}
\usepackage{multirow}

\title{Anomaly Detection in Automated Fibre Placement: Learning with Data Limitations}
\renewcommand{\shorttitle}{Anomaly Detection in AFP: Learning with Data Limitations}

\date{} 					

\newcommand{\rbox}[1]{\rotatebox{90}{#1}}
\newcommand{\orcid}{\includegraphics[scale=0.07]{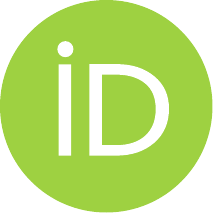} \hspace{1mm}}
\newcommand*\samethanks[1][\value{footnote}]{\footnotemark[#1]}

\makeatletter
\def\@fnsymbol#1{\ensuremath{\ifcase#1\or \dagger\or *\or \ddagger\or
   \mathsection\or \mathparagraph\or \|\or **\or \dagger\dagger
   \or \ddagger\ddagger \else\@ctrerr\fi}}
    \makeatother

\author{ \href{https://orcid.org/0000-0003-2961-5992}{\orcid Assef Ghamisi\thanks{These authors contributed equally to this work.}} \\
	Department of Electrical and Computer Engineering\\
	University of Victoria\\
	3800 Finnerty Rd.\\ Victoria BC, V8P 5C2, Canada\\
	\texttt{assefghamisi@uvic.ca}
 \And
	\href{https://orcid.org/0000-0001-5982-255X}{\orcid Todd Charter\samethanks} \\
	Department of Electrical and Computer Engineering\\
	University of Victoria\\
	3800 Finnerty Rd.\\ Victoria BC, V8P 5C2, Canada\\
	\texttt{toddch@uvic.ca}
 \And
	Li Ji \\
	LlamaZOO Interactive Inc.\\
	1019 Wharf St. \#200\\ Victoria BC, Canada\\
	\texttt{li.ji@llamazoo.com}
 \And
	Maxime Rivard \\
	National Research Council Canada\\
	1200 Montreal Road\\ Ottawa ON, K1A 0R6, Canada\\
	\texttt{maxime.rivard@cnrc-nrc.gc.ca}
 \And
	Gil Lund \\
    Fives Lund LLC\\
    13536 Beacon Coal Mine Rd S\\
    Seattle WA, 98178, USA\\
	\texttt{gil.lund@fivesgroup.com}
 \And
	\href{https://orcid.org/0000-0002-3550-225X}{\orcid Homayoun Najjaran\thanks{Corresponding Author}} \\
	Department of Mechanical Engineering \\
	University of Victoria\\
	3800 Finnerty Rd.\\ Victoria BC, V8P 5C2, Canada\\
	\texttt{najjaran@uvic.ca}
}

\hypersetup{
pdftitle = {Anomaly Detection in AFP: Learning with Data Limitations},
pdfsubject = {Anomaly Detection in Automated Fibre Placement: Learning with Data Limitations},
pdfauthor = {Assef Ghamisi, Todd Charter, Li Ji, Maxime Rivard, Gil Lund, Homayoun Najjaran},
pdfkeywords = {Automated Fibre Placement, Anomaly Detection, Computer Vision, Unsupervised Learning, Convolutional Autoencoder},
pdflang = {en-US},
colorlinks = true,
urlcolor = blue, 
linkcolor = blue, 
citecolor = red 
}

\begin{document}

\maketitle

\begin{abstract}
    Conventional defect detection systems in Automated Fibre Placement (AFP) typically rely on end-to-end supervised learning, necessitating a substantial number of labelled defective samples for effective training. However, the scarcity of such labelled data poses a challenge. To overcome this limitation, we present a comprehensive framework for defect detection and localization in Automated Fibre Placement. Our approach combines unsupervised deep learning and classical computer vision algorithms, eliminating the need for labelled data or manufacturing defect samples. It efficiently detects various surface issues while requiring fewer images of composite parts for training.
    Our framework employs an innovative sample extraction method leveraging AFP's inherent symmetry to expand the dataset. By inputting a depth map of the fibre layup surface, we extract local samples aligned with each composite strip (tow). These samples are processed through an autoencoder, trained on normal samples for precise reconstructions, highlighting anomalies through reconstruction errors. Aggregated values form an anomaly map for insightful visualization. The framework employs blob detection on this map to locate manufacturing defects.    
    The experimental findings reveal that despite training the autoencoder with a limited number of images, our proposed method exhibits satisfactory detection accuracy and accurately identifies defect locations. Our framework demonstrates comparable performance to existing methods, while also offering the advantage of detecting all types of anomalies without relying on an extensive labelled dataset of defects.
\end{abstract}


\section{Introduction}
\label{sec:intro}
Automated Fibre Placement (AFP) is an advanced composite manufacturing method for forming strong and lightweight components from strips of reinforced fibres known as tows. It is commonly used in quality-critical industries such as aerospace, where quality inspection and assurance are paramount \citep{palardy-sim_next_2019, bockl_effects_2023}. Most existing inspection techniques are implemented via manual human examination strip by strip, which is time-consuming, and thus a major production bottleneck \citep{meister_review_2021}. To address this problem, recent research seeks to automate defect detection by using artificial intelligence (AI), computer vision (CV) and deep learning (DL) methodologies, reducing manual effort in AFP inspection, and expediting the production \cite{brasington_automated_2021}.

As a widely adopted methodology, supervised learning with explicit labelling has been applied for inspection in AFP and similar industries. These methods leverage convolutional neural networks (CNNs) to train from extensive labelled datasets of surface imaging of manufactured parts. These surface images take a variety of forms, including photographs \cite{zemzemoglu_design_2022}, thermal images \cite{schmidt_deep_2019}, and depth maps from many types of profilometry sensors \cite{sacco_machine_2020, zhang_research_2022}. 

Sacco et al. (2020) \cite{sacco_machine_2020} review the applications of machine learning in composite manufacturing processes and present a case study of state-of-the-art inspection software for AFP processes. The presented inspection method uses a deep convolutional neural network for semantic segmentation to classify defects on a per-pixel basis. They use about 800 scans which is relatively a large dataset in this domain, yet the results show their method often misses some defects. Object detection is a well-developed subfield of computer vision in which models learn to recognize specific objects from a large dataset of labelled bounding boxes. Zhang et al. (2022) \cite{zhang_research_2022} offer an alternate approach using object detection. This work implements a modified YOLOv5 network, which is a popular and commercially available object detection model. With a large dataset of 3000 images containing five different defect types, their proposed model demonstrated effective performance in detecting those five defect types.

To achieve better real-time inspection, Meister et al. \cite{ meister_performance_2023} evaluate the use of convolutional and recurrent neural network architecture for analyzing laser-scanned surfaces line by line as 1D signals. The different network structures are assessed on both real and synthetic datasets, demonstrating sufficient performance. Through experimentation, the authors evaluate the effects of training and testing on differing data types (real or synthetic), realizing that deviations between the training and testing domain have a greater potential to impact the results of their proposed 1D analysis methodology. 

These supervised learning methods, however, can be impractical in industrial projects because they require large, unambiguously labelled training datasets which are not typically available. There are three key reasons for the lack of labelled training data. First, collecting real-world data from production machines is expensive and disruptive to existing production schedules. Second, defects and anomalies in real-world production are rare. To collect enough defect and anomaly samples for the models to learn from, one must collect a very large amount of data, adding to the training cost. Third, real-world defects can take many different forms, and there is no universally accepted standard of how a human inspector should delineate and record anomalies and defects, not to mention how to label them for machine learning \citep{heinecke_manufacturing-induced_2019}. Industrial practitioners of AFP manufacturing typically rely on organizational-specific standards and individual professional practices to identify and correct AFP anomalies. These separate standards and practices cannot be easily translated into a well-defined labelling strategy for labelling training sets. 

One solution is to create synthetic datasets which can be utilized for training supervised models. Using AI models for synthetic dataset creation has been explored in many applications with varying success. To address the limited defect data in AFP, \cite{meister_synthetic_2021} compares different data synthesis techniques for generating defect data useful to AFP inspection applications. The paper compares synthetic image datasets generated by various GAN-based models, even implementing a CNN-based defect classifier for analysis. However, there is a lack of comparison of the generated datasets to real-world data regarding image diversity and realism \cite{meister_synthetic_2021}. In a similar problem, for the task of machine fault detection where faulty data is scarce and normal data is abundant, \cite{ahang_synthesizing_2022} implemented a Conditional-GAN for generating fault data in different conditions from normal data samples. This paper provides a more in-depth analysis comparing generated data to real ground truth fault data, showcasing that generated feature distributions are similar to those of real faults. Such data generation approaches have demonstrated effectiveness, though they still require sufficiently large and representative datasets, and cannot generalize to unseen defects or anomalies.

Circumventing the need for large datasets with labelled defects, unsupervised anomaly detection methods focus on learning the high-level representations of non-defective, normal data to identify outlying anomalies. In AFP manufacturing, normal data is typically well-defined thanks to the simple, invariant structure of the tows (narrow strips of composite material) and limited layup manners used. In this study, we utilize the non-defective samples, which constitute the majority of any real-world AFP datasets, to train a classifier capable of discerning normal and abnormal composite structures.

Autoencoders, known for their capability of reconstructing input data, have emerged as potent tools for detecting anomalies within images \cite{albuquerque_filho_review_2022}. An autoencoder works by learning to encode normal input samples into a lower dimensional latent vector that can be decoded to reconstruct the original sample. The reconstruction is compared with the original, and a reconstruction error metric is calculated. When an abnormal sample is provided, the reconstruction errors will be high since the autoencoder was never trained with similar images. A threshold is then applied to the reconstruction errors to determine if the sample is normal, with a higher error indicating the sample is more likely to be abnormal or defective.

There is a lack of research applying these methods in the AFP industry, however, the approach has demonstrated success in other similar defect detection tasks. Ulger et al. (2021) \cite{ulger_anomaly_2021} employ convolutional and variational autoencoders (CAE and VAE) for solder joint defect detection. Reconstruction errors guide classification, applying a threshold to differentiate between normal and abnormal inputs. Tsai et al. (2021) \cite{tsai_autoencoder-based_2021} also employ CAE and VAE for textured surface defect detection, favouring CAE in a Receiver Operator Characteristic (ROC) analysis. The proposed anomaly detector is tested on various textured and patterned surface types, including wood, liquid crystal displays, and fibreglass.  Additionally, Chow et al. (2020) \cite{chow_anomaly_2020} implement a CAE for concrete defect detection, introducing a window-based approach for high-resolution images. Their window-based implementation enables pixel-wise anomaly maps to provide localization and contextual understanding of anomalies.

\begin{figure}
    \centering
    \begin{subfigure}[b]{0.4\textwidth}
         \includegraphics[width=\textwidth]{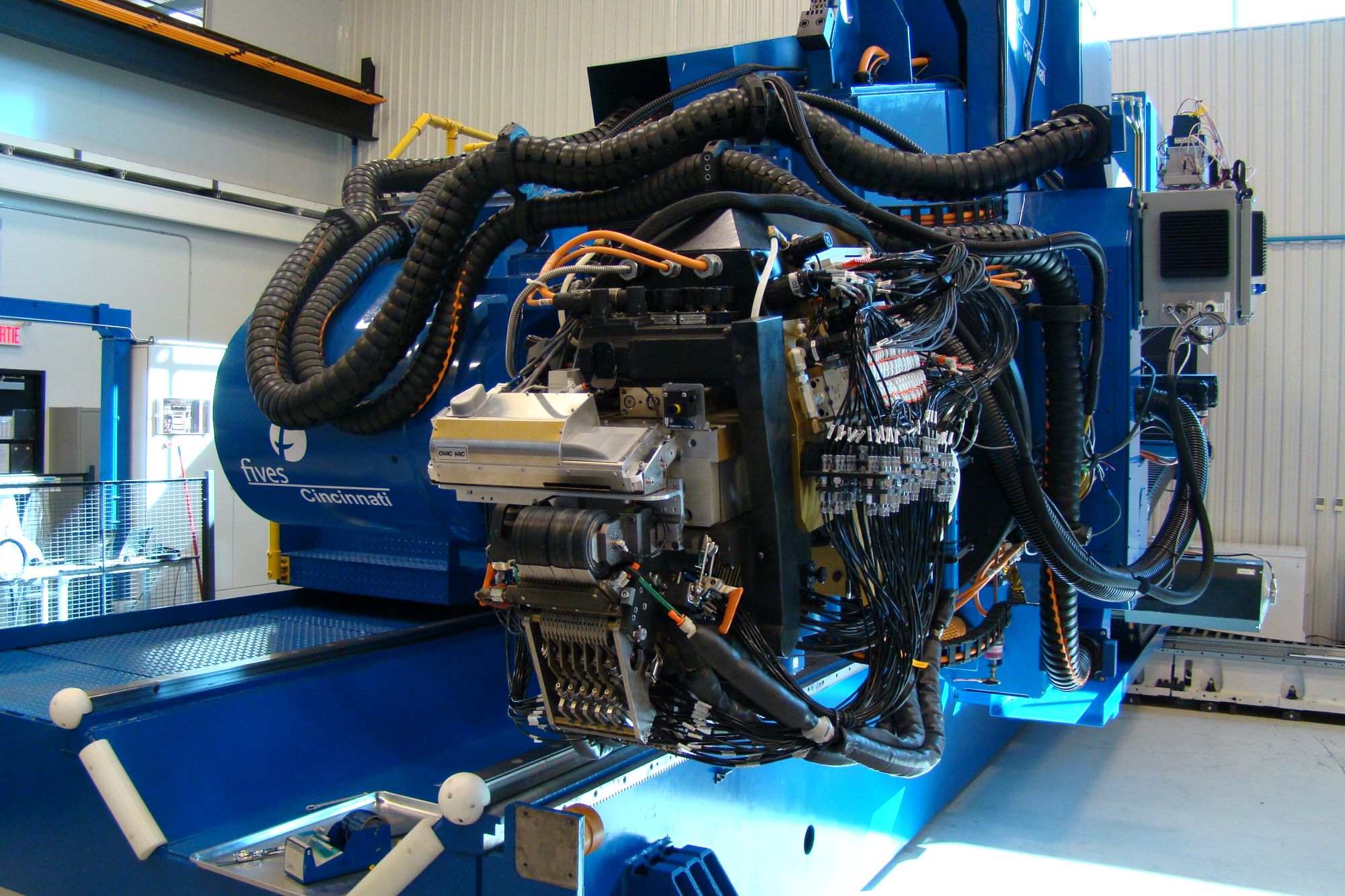}
         \caption{}
         \label{subfig:machine}
     \end{subfigure}
     \hfill
     \begin{subfigure}[b]{0.53\textwidth}
         \centering
         \includegraphics[width=\textwidth]{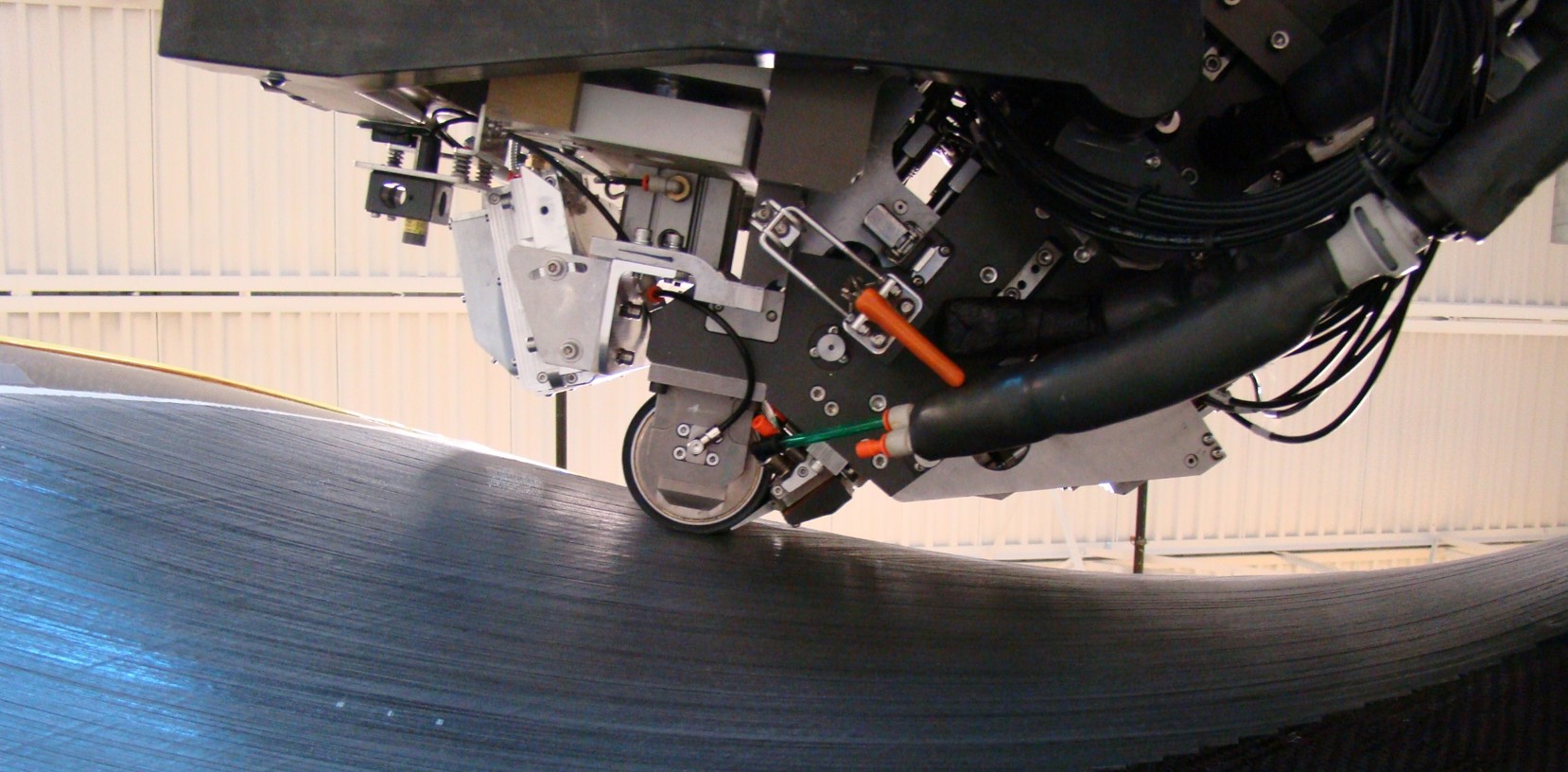}
         \caption{}
         \label{subfig:sensor}
     \end{subfigure}
        \caption{The industrial AFP setup is shown in two separate views. On the left is an overall view of the fibre placement machine (a), and on the right is a close-up shot of the robotic tool applying carbon fibre tows. The OCT sensor is visible to the upper left of the roller.}
        \label{fig:hardware}
\end{figure}

We propose a comprehensive framework for anomaly detection in AFP based on the autoencoder methodology. Compared to the existing methods that require a large labelled dataset including manufacturing defects, our approach is compatible with a small training dataset of normal samples. Our autoencoder-based anomaly detector uses data collected from the AFP setup shown in Figure \ref{subfig:machine}. The autoencoder is trained on a collection of local samples taken from depth images of the composite carbon fibre surfaces. The depth images are obtained with an Optical Coherence Tomography (OCT) sensor installed on the AFP layup head, which captures high-resolution point clouds of the layup tow surfaces \citep{rivard_enabling_2020, palardy-sim_advances_2019}. A picture of the AFP head with the OCT sensor installed on it is provided in Figure \ref{subfig:sensor}.

\begin{figure}
    \centering
    \begin{subfigure}[b]{0.55\textwidth}
         \includegraphics[width=\textwidth]{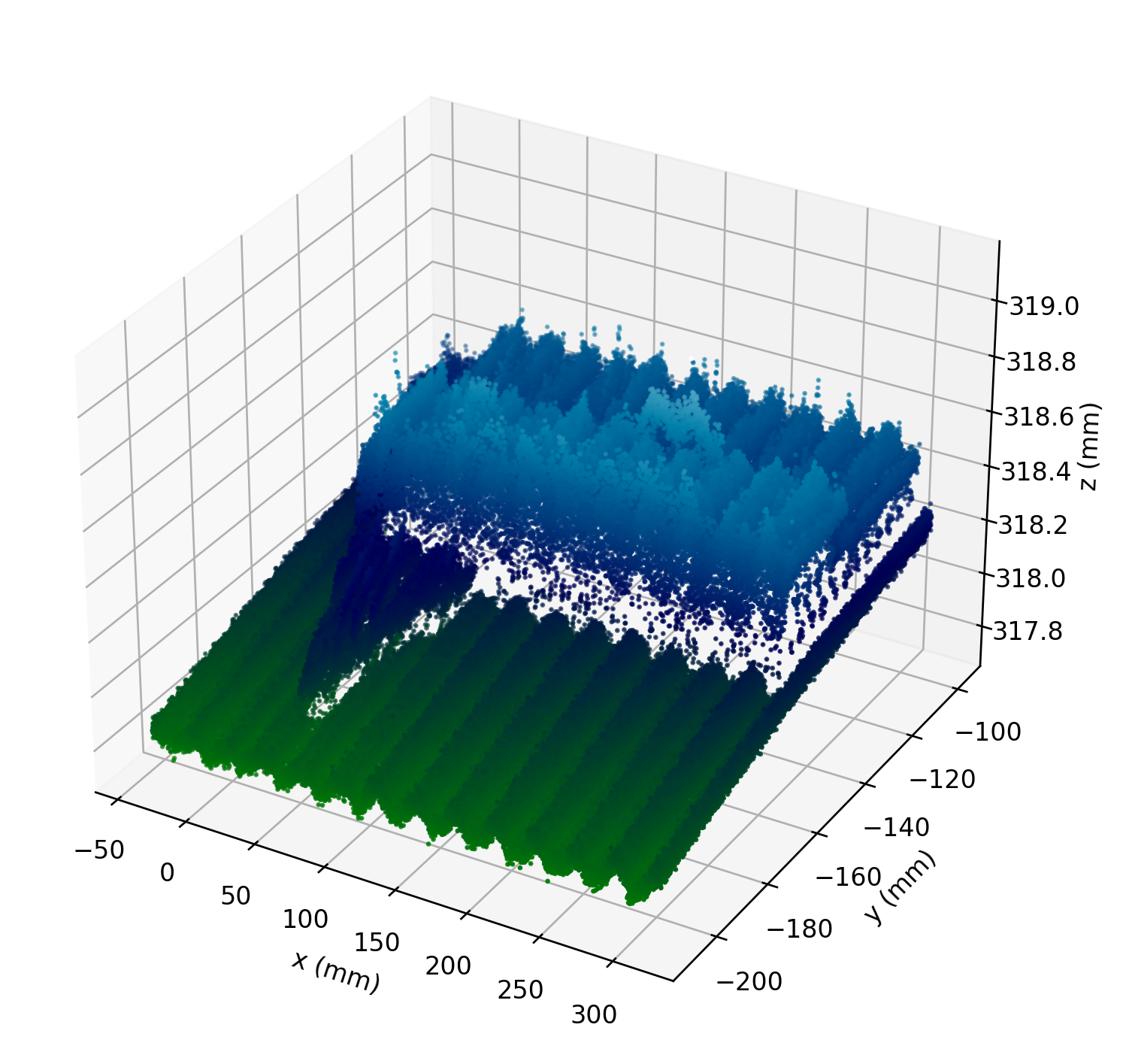}
         \caption{}
         \label{subfig:3dplot}
     \end{subfigure}
     \hspace{0.5\textwidth}
     \begin{subfigure}[b]{0.35\textwidth}
         \centering
         \includegraphics[width=\textwidth]{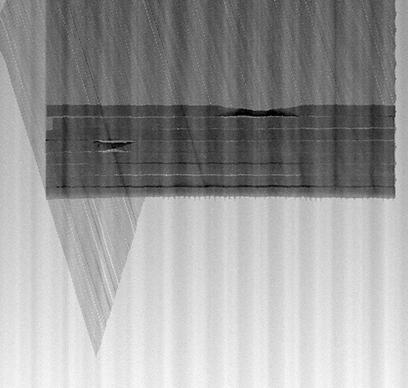}
         \caption{}
         \label{subfig:scan}
     \end{subfigure}
     \hspace{0.05\textwidth}
     \begin{subfigure}[b]{0.35\textwidth}
         \centering
         \includegraphics[width=\textwidth]{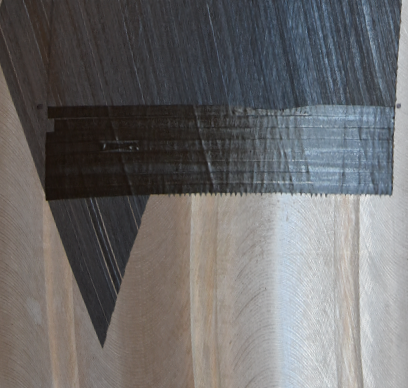}
         \caption{}
         \label{subfig:real}
     \end{subfigure}
        \caption{Different representations of a composite part manufactured with an AFP machine. Above (a) is a 3D point cloud measured using OCT Technology. The bottom left (b) shows the depth map generated from the 3D point cloud, and a real photograph of the same composite part is shown in the bottom right (c) for comparison.}
        \label{fig:composite_part}
\end{figure}

To simplify and enhance efficiency in processing information, the 3D point clouds are converted into 2D depth maps. Since the point cloud is a measurement of the surface elevation, it can be projected onto a 2D depth map without the loss of information.
These depth maps are presented as grayscale images, where the brightness of each pixel corresponds to the surface elevation on the composite part. Fig. \ref{fig:composite_part} provides various representations of a sample composite part, highlighting that defects are less discernible in the photograph due to reflections and low visual contrast. However, defects become more evident in the depth map, facilitating defect detection.

The work offers the following contributions.

\begin{enumerate}
    \item We introduce a novel, end-to-end framework for anomaly detection and localization in Automated Fibre Placement, circumventing the data limitations in this industry.
    Our proposed framework has several advantages compared to the existing methods. It detects all types of anomalies, without the need for manual data labeling or defect samples. Also, it works with a small number of composite images.
    \item Another major contribution of this work is an efficient data extraction methodology that can convert a limited number of composite images into a large dataset of local samples. Utilizing classical computer vision algorithms to detect the boundaries of composite tapes, this method generates a dataset that exploits the inherent symmetry of AFP composite materials.
    \item We design and validate an autoencoder with the optimal size of the latent domain that can identify the best distinctive features to differentiate between normal and defective samples.
    \item The proposed framework generates a map representing the local anomaly score of the AFP-manufacture parts and visualizes this map on the original composite scan. This visual representation serves as a valuable tool for AFP technicians, aiding them in the identification and resolution of anomalies within the composite structure.
\end{enumerate}

The organization of this manuscript is as follows: Section \ref{sec:methodology} describes, the whole procedure of anomaly detection, including data prepossessing, training of the AI model, and implementation details. In Section \ref{sec:results}, the evaluation results of both the anomaly detector and the localization system are presented. Finally, concluding remarks of this work are provided in Section \ref{sec:conclusion}.

\section{Methodology} \label{sec:methodology}
\begin{figure}
    \centering
    \includegraphics[width=0.9\linewidth]{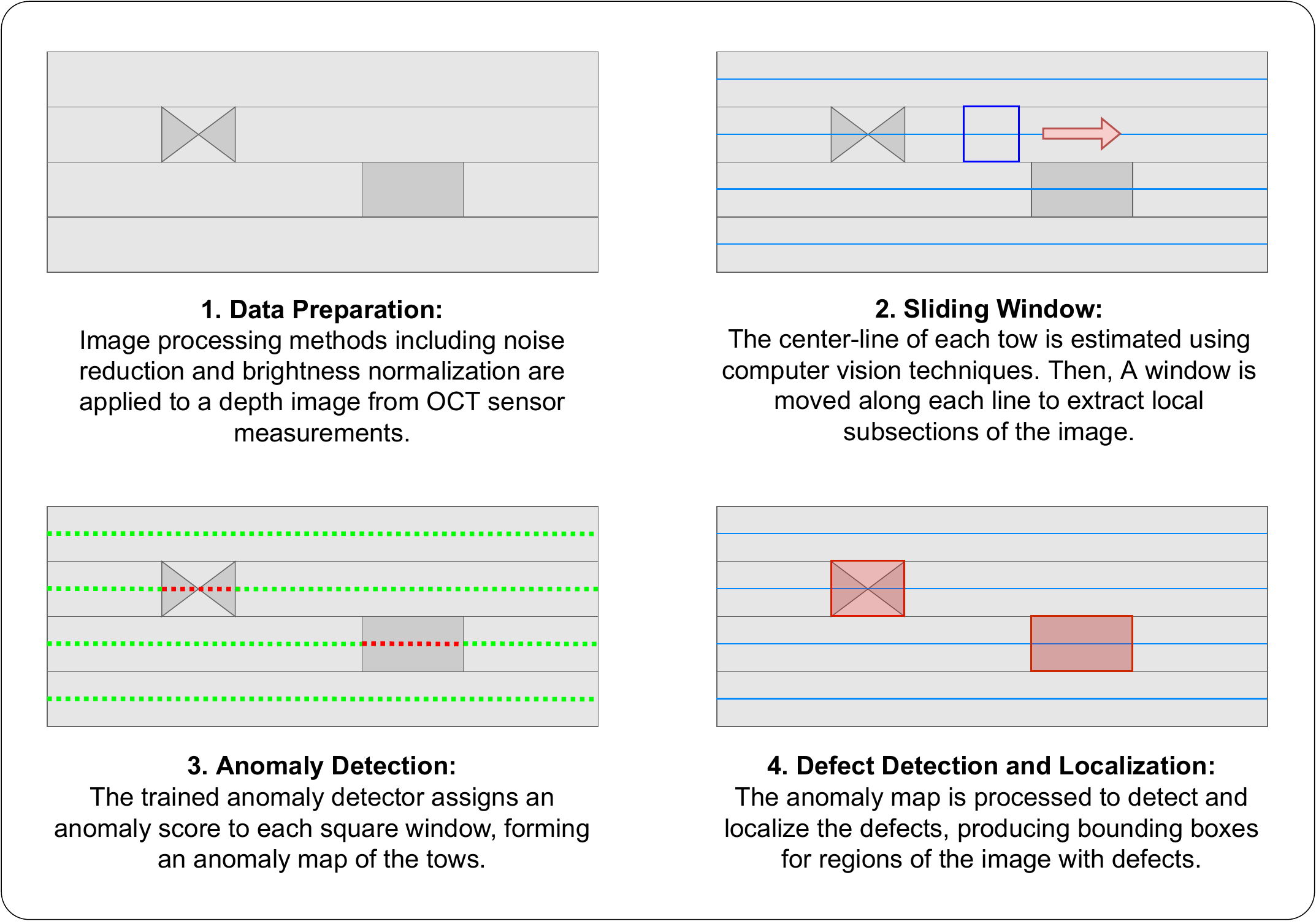}
    \caption{An overview of the defect detection process shows the necessary steps.}
    \label{fig:methodology}
\end{figure}

Figure \ref{fig:methodology} summarizes our anomaly detection framework. First, a composite scan is processed and local samples are extracted from the images. Then, the trained autoencoder generates an anomaly map, used to detect and locate the defects in the image.

\subsection{Data Preparation}

The raw depth maps contain impulse artifacts, also known as salt and pepper noise, which can be detrimental. To remove the noise, we used a median filter with the kernel size $3\times3$ applied to the whole depth image  \citep{azzeh_salt_2018}. Compared to a Gaussian filter with a small radius (a low-pass filter) \citep{deng_new_2016}, a median filter has less risk of losing high-frequency features.

Another data preparation step is needed because different raw depth-map images have inconsistent ranges of values depending on the distance of the laser origin to the composite surface. This effect is commonly caused when the OCT sensor is mounted to a fixed location behind the AFP head while scanning a contoured surface. These variations can cause undesired behaviour in the defect detection methodology. To address this, all the images undergo a min-max normalization so that the minimum depth value is mapped to zero and the maximum value is mapped to one. By applying this linear transformation, the visual contrast of the images is improved while keeping the original depth ratio. The normalization function is provided in Eq. \ref{eqn:min-max} in which $z_{i,j}$ is the original depth value, $p_{i,j}$ is the normalized pixel value, and $Z$ is the whole depth map matrix.
\begin{equation}
\label{eqn:min-max}
p_{i,j} = \frac{z_{i,j} - \min(Z)}{\max(Z) - \min(Z)}
\end{equation}

\subsection{Local Sample Extraction}

The training dataset used in this work is composed of depth maps from 42 non-defective composite surfaces. Developing an effective end-to-end network for defect detection using such a limited number of scans presents a significant challenge. However, we address this issue by leveraging the consistent uniformity along the composite tows and extracting cropped windows of the scans to form a dataset with many more localized samples. This is possible under the assumption that each cropped section conforms to a similar distribution, given that defect-free tows should exhibit minimal to no disparity across their segments. Consequently, the analysis of smaller regions allows us to employ a more compact neural network to learn from a broader spectrum of local samples rather than relying on a larger network to process full scans. Moreover, by extracting localized samples along the tows, our network gains exposure to a greater variation of tow structure.

\begin{figure}
    \centering
     \begin{subfigure}[b]{0.48\textwidth}
         \centering
         \includegraphics[width=\textwidth]{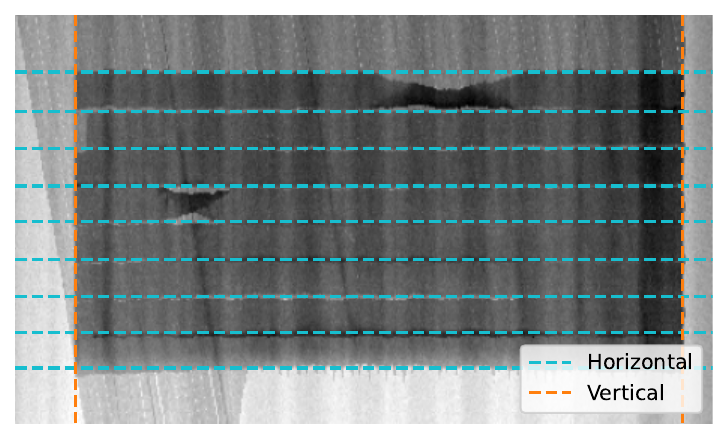}
         \caption{}
         \label{subfig:line}
     \end{subfigure}
     \hfill
     \begin{subfigure}[b]{0.48\textwidth}
         \centering
         \includegraphics[width=\textwidth]{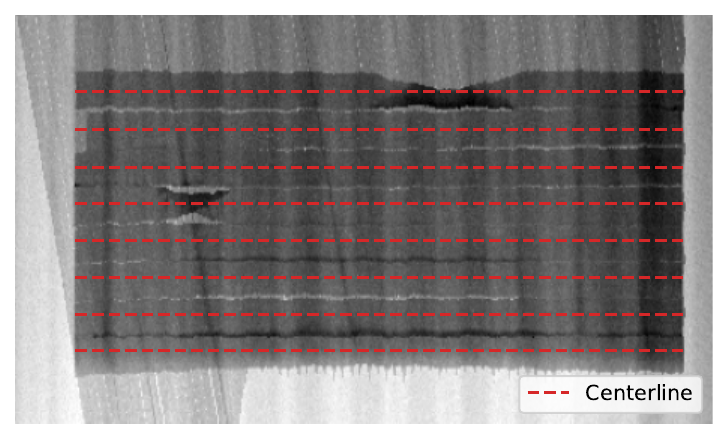}
         \caption{}
         \label{subfig:centerlines}
     \end{subfigure}
        \caption{The centerline detection procedure contains two main steps: detecting horizontal and vertical lines (a) and estimating tow centerlines from the detected lines (b).}
        \label{fig:line_detection}
\end{figure}

One of the most basic methods to detect local objects in an image is to move a window on the image and classify the smaller region inside the window \citep{dalal_histograms_2005}. This method is known as sliding window in computer vision literature and has its own limitations. For example, the scale of the object may vary depending on how close the object is to the camera. Consequently, multiple sizes of sliding windows are required which can be computationally expensive.
In our current dataset, on the other hand, most of the defects are localized to one tow, and therefore approximately the same relative size and there is no wide variation in perspective or orientation of the objects. Consequently, only one scale of sliding windows is sufficient for this use case. This also helps to keep computation complexity relatively low for this approach. Besides, there is preliminary knowledge of the composite part scans, like the number of tows and the general direction they follow. This enables a customized sliding window method that makes use of the known information. Moreover, the depth maps generated from our OCT scans have a specific structure. For example, all tows are placed straight and horizontal in the images, the number of tows is known, and their width is also known. To incorporate this predetermined knowledge, a line detector algorithm based on Hough Transform \citep{hough_machine_1959, duda_use_1972} detects the vertical and horizontal edges of the tows. After detecting the boundaries of the tows, the center of them (centerlines) are calculated by averaging each two consecutive horizontal lines, bounded within the detected vertical lines. This process of centerline detection is illustrated in Figure \ref{fig:line_detection}. Finding the center of the tows makes it possible to directly focus on the regions that are candidates for defect instead of scanning the whole image. In other words, it creates a skeleton that directs and constrains the region of interest. This can reduce additional effort on the classifier side.

\begin{figure}
    \centering
    \includegraphics[width=0.7\linewidth]{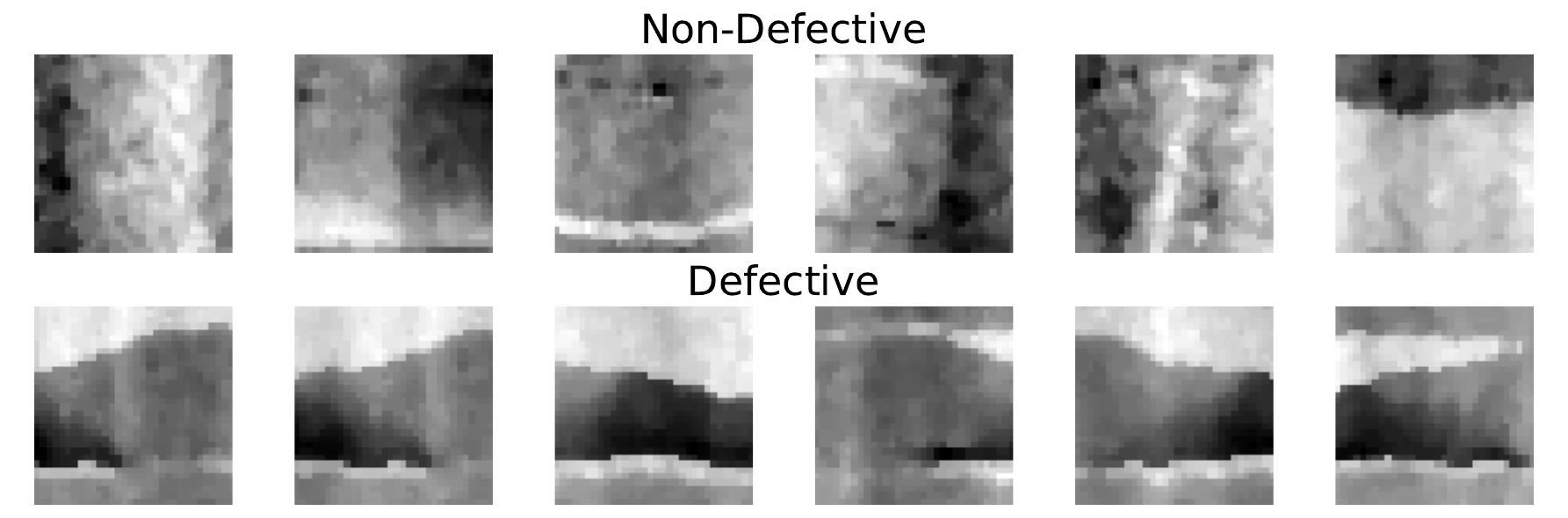}
    \caption{A dataset is created from cropped sections of the depth maps, using the sliding window method. Normal samples are shown on top and abnormal samples are shown below.}
    \label{fig:samples}
\end{figure}

Based on the detected centerlines, a square window slides across the tows to extract cropped regions. We select a window size of $32\times32$ pixels to cover approximately $1.5$ times the width of composite tapes. In this implementation, a stride of $8$ pixels is used to move the window and sample the information cropped inside. This combination of window size and stride allows enough overlap between the nearby samples while keeping the samples sufficiently distinctive. Some of the extracted samples are presented in Figure \ref{fig:samples}. At the time of inference, each window that is detected as an anomaly is considered to contain a defect, while the windows that are not anomalies are assumed to have normal tow structures. The next sections explain the approach to distinguishing between normal and abnormal samples.

\subsection{Anomaly Detection}

As mentioned in Section \ref{sec:intro}, autoencoders have shown great success at identifying anomalies in images. An autoencoder is an unsupervised learning model that reconstructs the given input by learning to minimize the error between the input and reconstructed output. They do this by encoding the input to a vector of latent features, also known as the bottleneck, and then decoding those latent features to reconstruct the input.
Convolutional Autoencoders (CAEs) are a group of autoencoders that use convolution layers in their network structure. Convolutional neural networks (CNN) are more popular for image-based autoencoders than basic fully-connected networks. This is because CNNs incorporate receptive fields using kernels that maintain the spatial relationships of the data. CNNs are also computationally efficient with sparse connectivity of neurons.

If only normal samples are used to train the autoencoder, it will be able to reconstruct similar normal samples accurately, and the reconstruction results for abnormal samples will be poor. Therefore, reconstruction error can be used as an indicator of how anomalous each input is. For inference, each cropped window of a composite material depth map is fed into the trained autoencoder. The reconstruction error of each window is then used as an anomaly score to create an anomaly map for the entire image. Reconstruction error of a window centred at $(x,y)$ is calculated using Eq. \ref{eqn:mse} in which $p_{i,j}$ and $\hat{p_{i,j}}$ are the pixel value of the input and reconstructed output, respectively, and $b$ is half of the size of each window.

\begin{equation}
\label{eqn:mse} 
m_{x,y} = \frac{1}{(2b)^2}\sum_{i=x-b}^{x+b-1}\sum_{j=y-b}^{y+b-1}[p_{i,j}-\hat{p}_{i,j}]^2
\end{equation}

\begin{figure}
    \centering
     \begin{subfigure}[b]{0.8\textwidth}
         \centering
         \includegraphics[width=\textwidth]{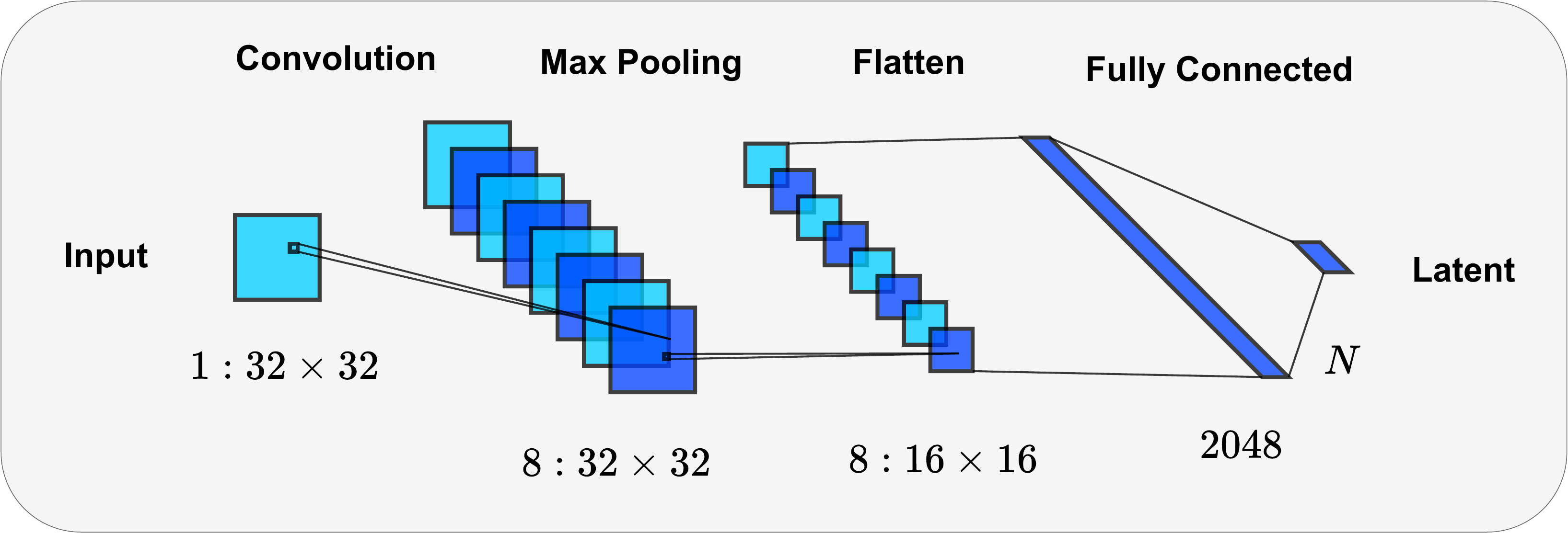}
         \caption{Encoder}
         \label{subfig:encoder}
     \end{subfigure}
     \hfill
     \begin{subfigure}[b]{0.8\textwidth}
         \centering
         \includegraphics[width=\textwidth]{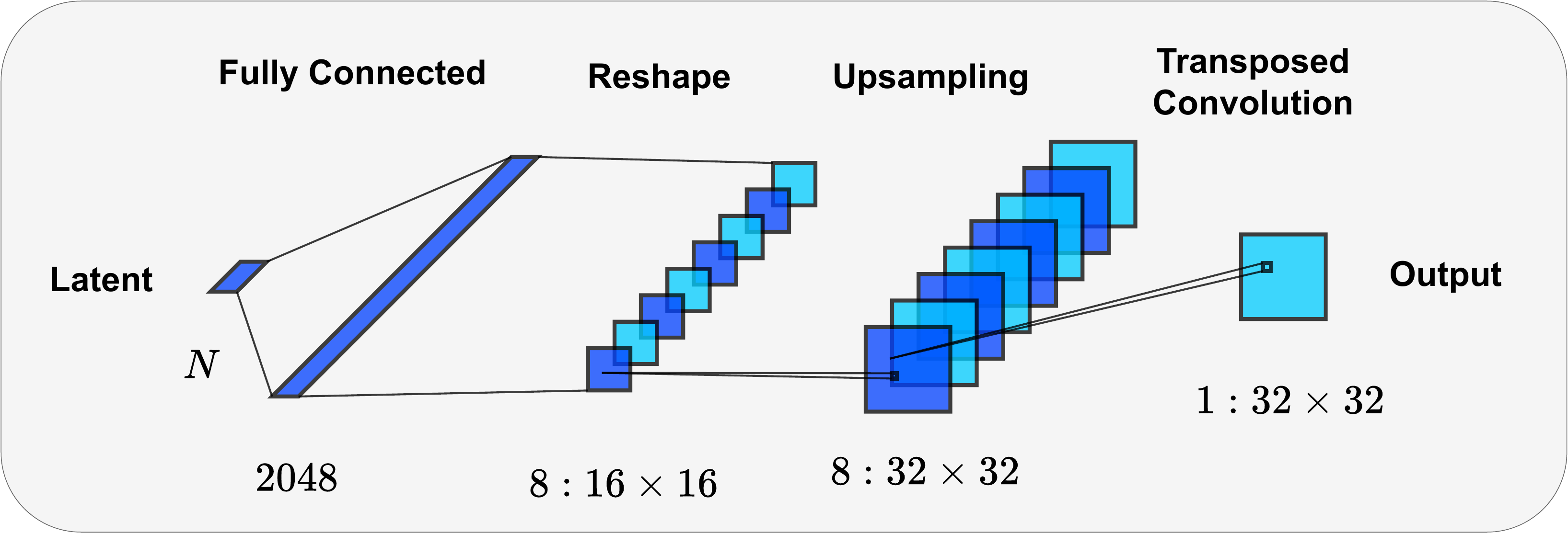}
         \caption{Decoder}
         \label{subfig:decoder}
     \end{subfigure}
        \caption{A graphic depicting the network structure of the proposed autoencoder. Above is the encoder structure (a), and below is the decoder structure (b).}
        \label{fig:network}
\end{figure}

In this work, a CAE is designed and used as the anomaly detector. The design incorporates symmetric encoder and decoder structures shown in figures \ref{subfig:encoder} and \ref{subfig:decoder} respectively. For training the model, mean squared error is employed as the loss function. Although the full method uses continuous valued anomaly maps to identify the defects rather than a binary prediction, binary classification can still be useful in validating the model performance. For this, a threshold parameter is introduced to classify samples based on their reconstruction error. To select the threshold value, a Receiver Operating Characteristic (ROC) curve is applied. The ROC curve plots the true positive rate against the false positive rate while varying the threshold value. In an ideal case, the selected threshold would give a true positive rate of 1 and a false positive rate of 0. In the ROC plot, this corresponds to the upper left corner and hence the best threshold value is selected from the curve at the point closest to that corner.

\subsection{Defect Localization}
The anomaly detection generates an array of anomaly scores for each tow, which can be considered as a 1D digital signal. Any area of this signal with a concentration of high values indicates the presence of a defect. In Computer Vision literature, these areas are called blobs \citep{danker_blob_1981, kong_generalized_2013}. For detecting the blobs, we use the Difference of Gaussian (DoG) method \citep{lowe_distinctive_2004}. In this approach, the signal ($f(x)$) is filtered using Gaussian kernels with increasing values for standard deviations ($\sigma$) as described in Eq. \ref{eqn:blob}. Then, the subtractions of each two successively filtered signals  are calculated. The local maxima of $g(x,\sigma)$ represents the blobs. In such maxima points, $x$ and $\sigma$ correspond to the location and characteristic scale (size) of the blob, respectively.

\begin{equation}
\label{eqn:blob}
g(\sigma,x) = {\sigma}^2\frac{\partial n_{\sigma}^2}{\partial x^2} \ast f(x)
\end{equation}

For each defect, two parameters are detected, radius and center. With this information the detected blobs can be transferred from anomaly map to image space.

\section{Results and Discussion} \label{sec:results}
The performance of the anomaly detection and localization system depends on two factors. First is the number of samples the anomaly detector is correctly classifying as normal or abnormal. Second is the size and location accuracy of the predicted defects. This section evaluates these two aspects using a test dataset with an additional two composite surfaces containing defects.

\subsection{Anomaly Detection}

The network structure proposed in Figure \ref{fig:network} is implemented using three different latent dimensions 2, 16, and 128 for comparison. Each network is trained with a dataset consisting of 27406 only normal samples. An Adam optimizer is employed with an MSE loss function to train the network. The batch size is set to 128. Each autoencoder undergoes training for 50 epochs, completing in under 5 minutes on a computer with the following specifications:
\begin{itemize}
    \item \textbf{Processor (CPU):} Intel(R) Xeon(R) E5-1607 v4 @ 3.10GHz
    \item \textbf{Graphics (GPU):} NVIDIA GeForce GTX 1080
    \item \textbf{Memory (RAM):} 32.0 GB
\end{itemize}

\begin{figure}
    \centering
    \includegraphics[width=0.5\linewidth]{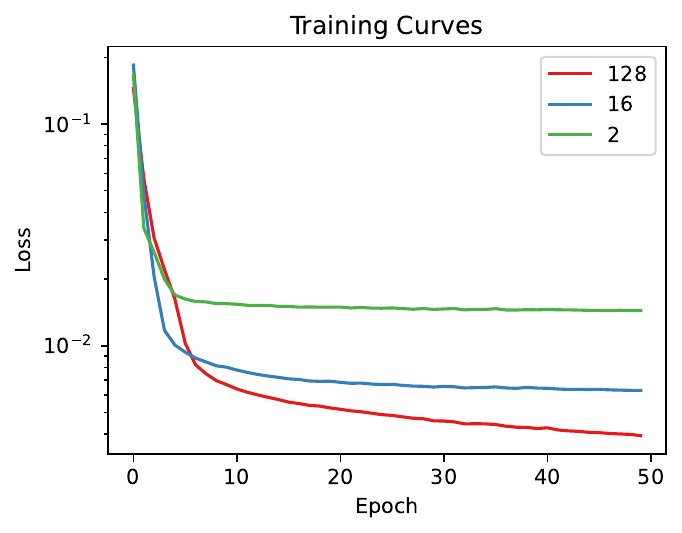}
    \caption{The training MSE losses of the three autoencoders are plotted in comparison over 50 epochs.}
    \label{fig:ae_train}
\end{figure}

The curves in Figure \ref{fig:ae_train} demonstrate the training losses of each autoencoder. Comparing the training loss curves shows that the models' reconstruction ability improves with higher dimensional latent space. The curves also show that the models are learning relatively quickly with tiny improvement in the later epochs.

\begin{figure}
\centering
\includegraphics[width=0.6\textwidth]{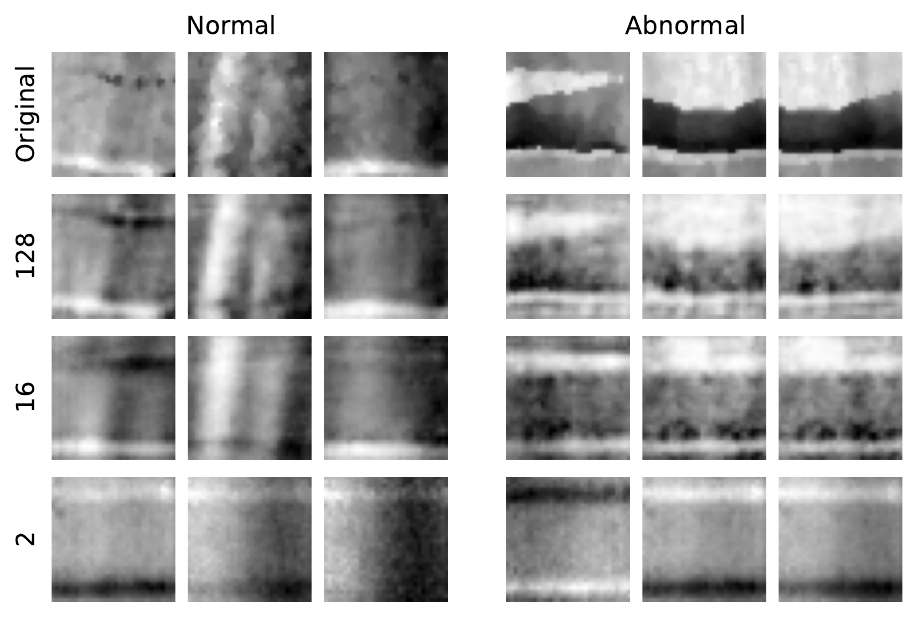}
\caption{The resulting reconstructions from the autoencoders with various latent sizes are compared for both normal and abnormal test samples.}
\label{fig:reconstructions}
\end{figure}

Reconstruction results for the autoencoders are demonstrated in Figure \ref{fig:reconstructions}. The original samples are randomly selected from normal and abnormal classes in test set.
These results clearly show the improved reconstruction performance with a higher dimensional latent space. It shows that the autoencoder with a 128-dimensional latent vector is able to produce good reconstructions for both normal and abnormal samples. The 16-dimensional autoencoder, on the other hand, produces relatively good reconstructions for normal samples and poorer reconstructions of defect samples. This is ideal for the classification method to distinguish anomalies. Finally, the autoencoder with only a 2-dimensional latent space is unable to make good reconstructions for any of the input samples.

\begin{figure}
     \begin{subfigure}[b]{0.48\textwidth}
         \centering
         \includegraphics[width=\textwidth]{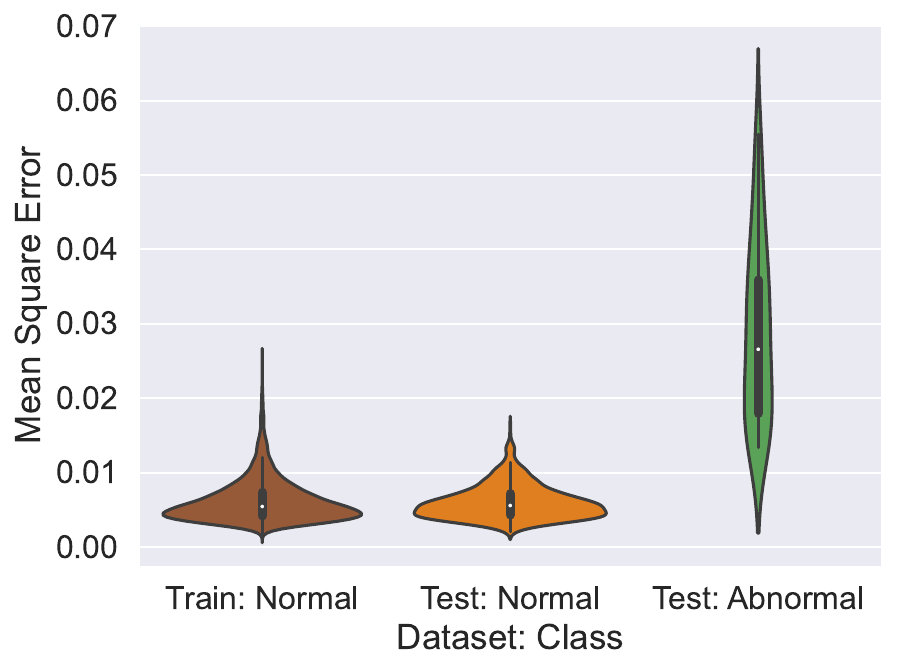}
         \caption{Comparing Test and Training}
         \label{subfig:violinplot}
     \end{subfigure}
     \hfill
     \begin{subfigure}[b]{0.48\textwidth}
         \centering
         \includegraphics[width=\textwidth]{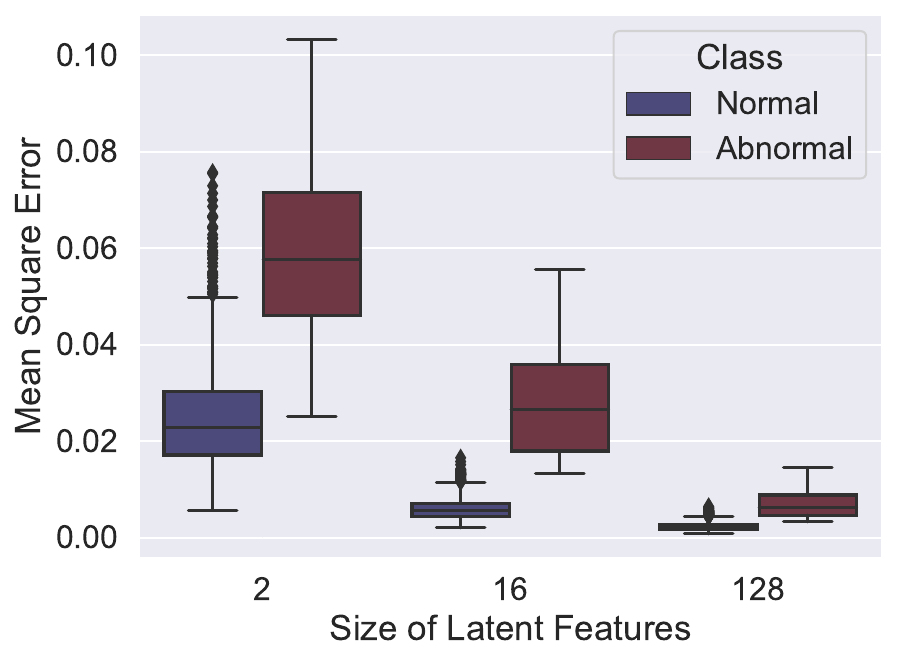}
         \caption{Comparing Different Latent Sizes}
         \label{subfig:boxplot}
     \end{subfigure}
        \caption{Distributions of MSE for different input types and latent sizes are presented for comparison.}
        \label{fig:mse}
\end{figure}

Figure \ref{fig:mse} shows comparisons of the reconstruction error. In figure \ref{subfig:violinplot} the distributions of mean square error are shown for the 16-dimensional autoencoder on the training set and test set. Note that the training set only includes normal samples whereas the test set contains both normal and abnormal samples, separated accordingly. As the figure suggests, the normal samples have a similar distribution in both the training and test sets. On the other hand, the abnormal samples have a generally higher MSE with a slight overlap on normal sample distribution. In an ideal case, if there were no overlap between these two distributions we could find a perfect threshold as the decision boundary to classify the samples into normal and abnormal categories. With existing overlap, however, an ROC curve can help to select the decision boundary that makes the best trade-off between true and false positive rates. Taking a closer look at the difference between the three autoencoders, Figure \ref{subfig:boxplot} shows boxplots of the MSE for normal and abnormal samples in the test set, using different number of latent features. Here it shows how separable the two classes are based on reconstruction error alone. For the 2-dimensional autoencoder, the interquartile ranges are separable, but there is a significant overlap when considering the whiskers. For the 16-dimensional autoencoder, the separation is greatly improved with minimal overlap between the whiskers. The 128-dimensional autoencoder, however, does not show significant separation and would be impossible to accurately classify the two classes based on MSE alone.

\begin{figure}
    \centering
    \includegraphics[width=0.5\linewidth]{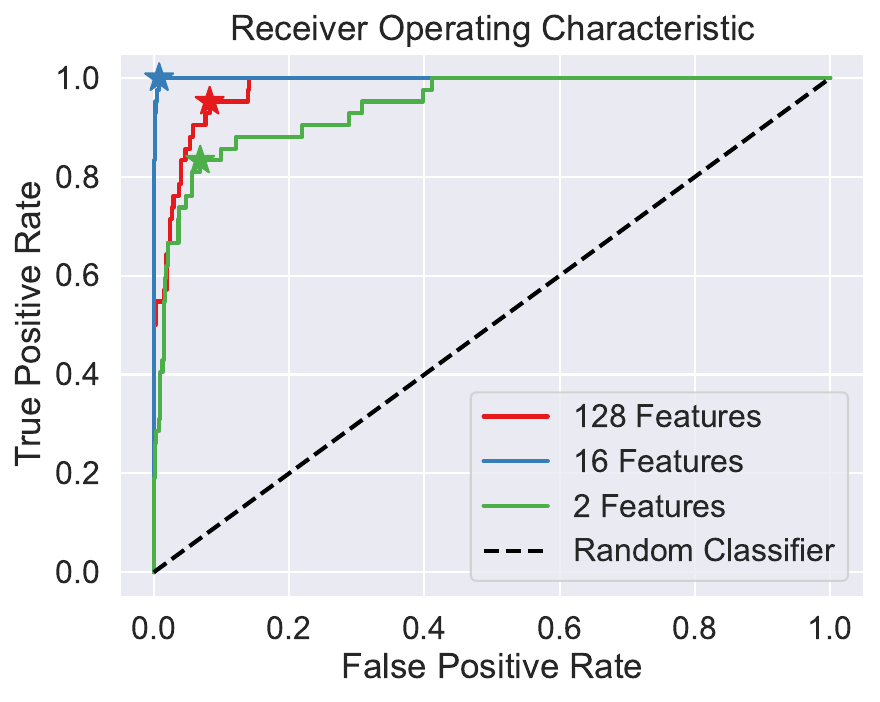}
    \caption{ROC curves of the test set are plotted for the three autoencoder classifiers.}
    \label{fig:roc_curves}
\end{figure}

Figure \ref{fig:roc_curves} shows the ROC curves for each of the three autoencoders and the selected best threshold points shown as stars. These results further demonstrate that classification performance does not correspond with reconstruction performance, as the 2D and 128D autoencoders' ROC curves have worse classification performance than the 16D model. The best-performing model is the autoencoder with a 16-dimensional latent space, achieving a high true positive rate with a low false positive rate.

Classification results of each autoencoder with the selected threshold values found by the ROC curves are summarised in Table \ref{table:ae_metrics}. The table reports precision, recall, F1 score, area under the ROC curve (AUC), and selected threshold for each classifier.

\begin{table}
\caption{Test Metrics of Convolutional Autoencoder}
\begin{center}
\begin{tabular}{cccccc}
\hline
Latent Dim. & AUC & Best Threshold & Precision & Recall & F1-Score \\
\hline
128 & 0.978 & 3.86 & 0.918 & 0.929 & 0.923\\
16 & 0.999 & 13.69 & 0.992 & 0.976 & 0.984\\
2 & 0.942 & 42.08 & 0.922 & 0.810 & 0.862\\


\hline
\end{tabular}
\label{table:ae_metrics}
\end{center}
\end{table}

\begin{figure}
     \begin{subfigure}[b]{0.48\textwidth}
         \centering
         \includegraphics[width=\textwidth]{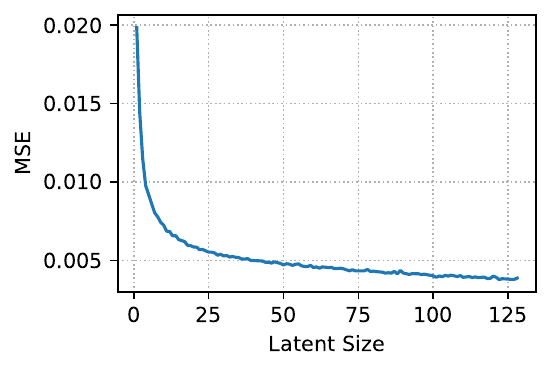}
         \caption{MSE}
         \label{subfig:lowest_loss}
     \end{subfigure}
     \hfill
     \begin{subfigure}[b]{0.48\textwidth}
         \centering
         \includegraphics[width=\textwidth]{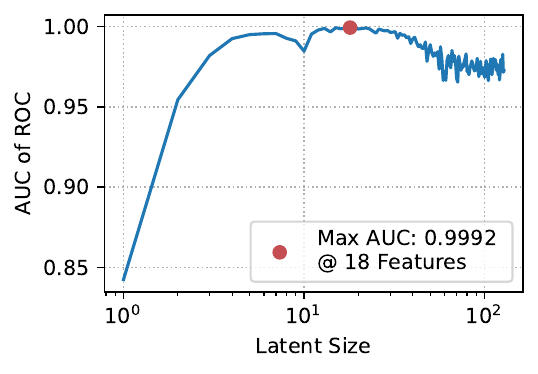}
         \caption{AUC of ROC}
         \label{subfig:auc_all}
     \end{subfigure}
        \caption{MSE and AUC of ROC are plotted for autoencoder models with latent dimensions varied from 1 to 128.}
        \label{fig:best_latent}
\end{figure}

Figure \ref{fig:best_latent} shows the effect of the latent vector size on the performance of the model. Figure \ref{subfig:lowest_loss} suggests that larger latent dimensions will produce lower reconstruction errors. However, an accurate classification model does not require the lowest reconstruction error, but a moderate reconstruction that leads to better classification performance. In Figure \ref{subfig:auc_all} the best latent dim is found by calculating the minimum AUC of the ROC curve while varying the latent dim.

\begin{figure}
    \centering
    \includegraphics[width=0.4\linewidth]{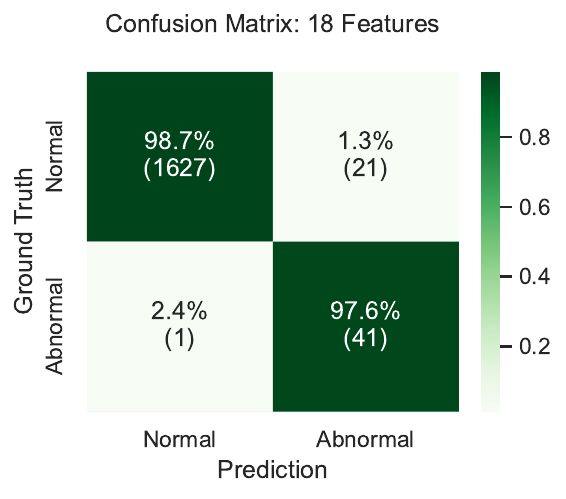}
    \caption{The confusion matrix demonstrates the evaluation of the best model.}
    \label{fig:confusion_matrix}
\end{figure}

Figure \ref{fig:confusion_matrix} shows the classification confusion matrix while using the optimal latent size. The off-diagonal values in this matrix are low which shows that most of the samples from both normal and abnormal classes are classified correctly.

\subsection{Defect Localization}

\begin{figure}
    \centering
    \includegraphics[width=0.8\linewidth]{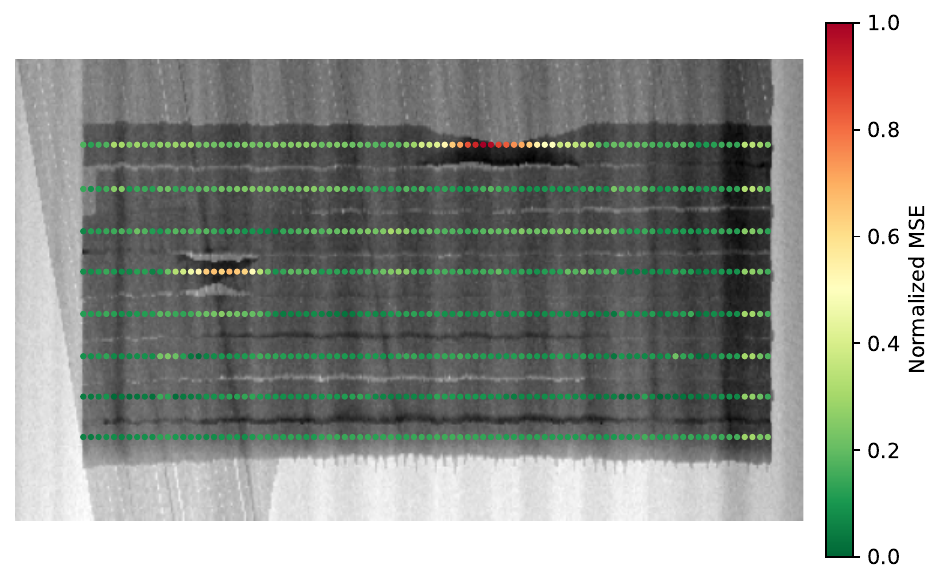}
    \caption{An anomaly map is generated from the MSE of individual cropped windows.}
    \label{fig:anomaly_map}
\end{figure}

Figure \ref{fig:anomaly_map} illustrates the results of the anomaly detector on a 2D depth map. The colour of each point indicates the normalized MSE for reconstructing a small window around the point with the anomaly detector. As can be seen, the defective areas have a large density of points with a higher MSE. This information can be used to detect these areas while ignoring individual outlier values.

\begin{figure}
    \centering
    \includegraphics[width=0.6\linewidth]{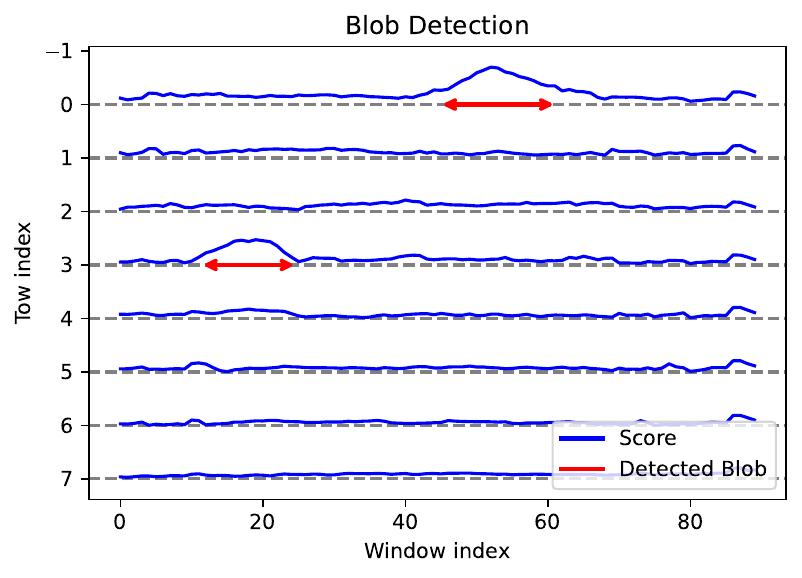}
    \caption{Anomaly scores are visualized as 1D signals for blob detection.}
    \label{fig:blobs}
\end{figure}

In Figure \ref{fig:blobs} the process of detecting the defects from the anomaly map is illustrated. The elevation in each curve represents the MSE value for one tow (represented by colour in the previous figure). The arrows show the detected blobs after applying the Derivative of Gaussians method. It is observed that only the areas with an extended length of high MSE values are detected as blobs.

\begin{figure}
    \centering
    \includegraphics[width=0.6\linewidth]{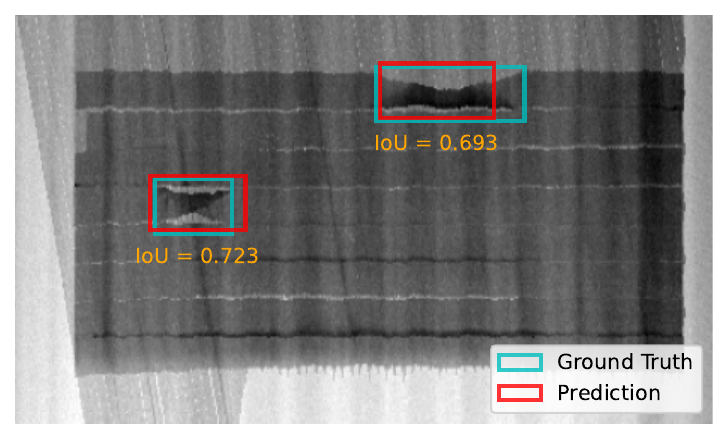}
    \caption{Predicted bounding boxes are displayed on the original depth map in comparison with the ground truth bounding boxes. Also, the values of Intersection over Union are displayed.}
    \label{fig:bboxes}
\end{figure}

Figure \ref{fig:bboxes} shows the final output of the computer vision pipeline, comparing the annotated defect bounding boxes (ground truth) with the predicted bounding boxes.

\subsection{Qualitative Comparison}
The proposed framework is unique and to the best of our knowledge, no other studies have implemented an end-to-end unsupervised defect detection method for AFP inspection. Unfortunately, there are no publicly available datasets in this domain to serve as a benchmark for AFP inspection tasks. Additionally, the dataset used in this work is insufficient for training supervised learning models, which constitute the majority of current studies in this field. For these reasons, an explicit quantitative comparison of our method with other state-of-the-art approaches is not possible. However, a qualitative comparison of the most relevant studies is presented in Table \ref{table:comparison}, showcasing the advantages of our framework.

In regard to other defect detection methods, the main advantages of our proposed approach stem from the unsupervised learning process which enables learning with data limitations.
Foremost, our method detects all types of surface anomalies in AFP, whereas existing methods are limited to specific defect types. Additionally, unlike other methods, ours does not require labelling which is time-consuming and prone to errors. Moreover, our proposed framework works with fewer composite scans and it does not need any samples of defects.

Some methods implement semantic segmentation which requires explicit pixel-wise labelling. This is not necessarily needed, as our method provides sufficient localization with bounding boxes and anomaly maps.
Unlike other methods, ours does not classify the defects, however, a separate classification module could easily be integrated using the detected bounding boxes. Besides, in an industry where the majority of the inspected parts are non-defective, directly detecting the defective parts can reduce most of the effort.

\begin{table}
\centering
\caption{A Comparison of Existing Learning-based AFP Defect Detection Methods with the Proposed Framework} 
\footnotesize
\footnotesize
\begin{tabular}{@{}lllllllllll@{}}
\toprule
\multirow{2}{*}{Reference} & \rbox{\multirow{2}{*}{Detection}} & \rbox{\multirow{2}{*}{Classification}} & \rbox{\multirow{2}{*}{Localization}} & \rbox{\multirow{2}{*}{Segmentation}} & \rbox{\multirow{2}{*}{Detects All Abnormalities}} & \rbox{\multirow{2}{*}{Labelling not Required}} & \rbox{\multirow{2}{*}{Defects not Required}} & \rbox{\multirow{2}{*}{Dataset Size}} & \multicolumn{2}{c}{Results} \\ \cmidrule(l){10-11} 
 &  &  &  &  &  &  &  &  & Classification & Localization \\ \midrule

Meister and Wermes (2023) \cite{meister_performance_2023} & \checkmark & \checkmark & - & - & -  & - & - & 469 & Acc$^{1}$: 95.12\% & - \\
Zhang et al. (2022) \cite{zhang_research_2022}      & \checkmark & \checkmark & \checkmark & - & -  & - & - & 3,000 & - & mAP$^{4}$: 93.1\% \\
Tang et al. (2022) \cite{tang_-process_2022}       & \checkmark & \checkmark & \checkmark & \checkmark & - & - & - & 43 & Pr$^{2}$: 84.4\%, Re$^{3}$: 77.1\% & mIoU$^{5}$: 0.776\\
Sacco et al. (2020) \cite{sacco_machine_2020}       & \checkmark & \checkmark & \checkmark & \checkmark & -  & - & - & 800 & Average Acc$^{}$: 49.1\%  & - \\
Schmidt et al. (2019) \cite{schmidt_deep_2019}        & \checkmark & \checkmark & - & - & -  & - & - & 12,000 & Acc$^{}$: >92\% & - \\
This work                       & \checkmark & - & \checkmark & - & \checkmark & \checkmark  & \checkmark & 44 & Acc$^{}$: 98.7\% & mIoU$^{}$: 0.708 \\ \bottomrule
 
\multicolumn{11}{r}{\scriptsize{1: Accuracy, 2: Precision, 3: Recall, 4: mean Average Precision, 5: mean Intersection over Union}}
\label{table:comparison}
\end{tabular}
\end{table}



\section{Conclusions} \label{sec:conclusion}

This paper introduces a practical and novel method for the inspection of composite materials manufactured by Automated Fibre Placement (AFP). The AFP process is susceptible to various types of defects which can significantly impact the final product's quality, necessitating thorough inspection of the composite parts. Manual human inspection has traditionally been employed for this purpose, but it is time-consuming, labour-intensive, and prone to human errors. To enhance the efficiency, accuracy and reliability of AFP, the development of an automated inspection system is crucial. Current inspection procedures mostly utilize profilometry technologies like laser scanning, thermal imaging, and optical sensors to generate visual measurements of the part's surface. The data used in this work is obtained from a laser scanner that operates based on OCT technology, though the framework presented is general and can be adapted to other types of profilometry data.

In AFP inspection, robust and generalized supervised learning methods are infeasible due to limitations in available labelled data. Anomaly detection methods, on the other hand, can circumvent this challenge by focusing on learning the structure of normal samples to identify any abnormalities. The proposed computer vision framework detects individual tows in AFP composites and creates a dataset of sub-images by sliding a window along the center of each tow. The extracted data is then used to train an autoencoder designed to detect anomalies. Using the same sliding window procedure, the autoencoder produces anomaly scores for local regions of the composite part. These scores are aggregated to form an anomaly map of the full image. This anomaly map can then be used as an explicit indication tool by an operator. We further process this anomaly map using a 1D blob detection algorithm to generate bounding boxes around defects.

Compared to other state-of-the-art automated inspection methods in AFP, our approach offers several advantages. Since the autoencoder learns the inherent structure of normal tows, it is capable of detecting all anomalies, unlike other methods which can only detect defects specific to their training data. 
Furthermore, our suggested framework operates with fewer composite scans and eliminates the requirement for defect samples.
For verification purposes the autoencoder is evaluated in a classification task, achieving over 98\% classification accuracy. Additionally, the overall framework is implemented on a set of test samples where bounding boxes generated by the method achieve an Intersection over Union (IoU) of 0.708. This demonstrates sufficient accuracy in the localization of detected defects.

This paper outlines a novel defect detection approach for AFP and emphasizes its practicality, particularly in addressing data limitations. There are several potential directions for future research to enhance the inspection system's capabilities and extend its relevance to broader domains.
To improve dataset quality and quantity, we suggest investigating data engineering techniques like data augmentation and synthetic data generation.
Additionally, while the current system identifies anomalies, it lacks the ability to classify specific defect types. To address this, we recommend utilizing the generated bounding boxes to collect training data for developing a classification model. 
We also recommend adapting our framework for use in industries that share similar tape-by-tape structures. This can lead to enhancements in defect identification and quality assurance across different sectors.
\section*{Statements and Declarations}

\subsection*{Acknowledgements}
We would like to acknowledge the financial support of LlamaZOO Interactive Inc.
and Natural Sciences and Engineering Research Council (NSERC) Canada under the Alliance Grant ALLRP 567583 - 21. In addition, we would like to recognize the research collaboration of LlamaZOO Interactive Inc., the National Research Council of Canada (NRC), and Fives Lund LLC.

\subsection*{Conflict of Interest}
The authors declare that they have no conflicting interests or financial motives related to the work in this article.

\subsection*{Data Availability}
The data generated and analyzed in this paper is proprietary and due to its commercialization potential, is not made publicly available. The information is protected to maintain its commercial value. We regret any inconvenience this may cause and appreciate your understanding of the importance of preserving its proprietary nature.

\bibliographystyle{ieeetr}
\bibliography{main}

\end{document}